\documentclass[letterpaper]{article}

\usepackage[preprint]{aaai2027}
\usepackage[hyphens]{url}
\usepackage{graphicx}
\usepackage{natbib}
\usepackage{caption}
\usepackage{booktabs}
\usepackage{amsmath}
\usepackage{amssymb}

\title{Watch-Think-Interact: Bootstrapping Long-Horizon Multi-Turn Streaming Video Reasoning with Reinforcement Learning}

\author{
Ziheng Huang\textsuperscript{1}\equalcontrib,
Yicheng Bao\textsuperscript{1}\equalcontrib,
Xueheng Li\textsuperscript{2}\equalcontrib,
Zhenkun Gao\textsuperscript{1}\equalcontrib, \\
Bangwei Liu\textsuperscript{1}\equalcontrib,
Kunquan Li\textsuperscript{3},
Yuxiang Shen\textsuperscript{3},
Bangyan Li\textsuperscript{1}, \\
Xuejiao Wang\textsuperscript{1},
Changbo Wang\textsuperscript{1},
Gaoqi He\textsuperscript{1}\corresponding
}
\affiliations{
\textsuperscript{1}East China Normal University \quad
\textsuperscript{2}University of Science and Technology of China \quad
\textsuperscript{3}Xiamen University \\
51275901129@stu.ecnu.edu.cn \quad gqhe@cs.ecnu.edu.cn
}

\begin{document}
\maketitle

\begin{abstract}
Multi-turn streaming video reasoning must answer asynchronous questions from observed prefixes under a bounded online state.
Existing methods form this state before future questions are known, making details discarded by forward-only compression irrecoverable when their relevance emerges later.
We introduce Watch-Think-Interact (WTI), a unified approach that preserves on-demand access to observed source evidence under a bounded online state.
Our solution comprises a closed-loop inference framework, causally aligned interaction data, and trajectory-level optimization.
At inference, WTI couples compact natural-language memory with source-video time ranges, supporting direct reasoning and selective visual recall.
Its policy answers when evidence suffices, waits when evidence has not appeared, or recalls an observed interval and redecides at the same time step without replaying the full history.
WTI-82K organizes 82,335 timed questions into 4,812 trajectories aligning query times, answerable moments, evidence, state updates, and actions.
It supervises when to answer, wait, recall, and update memory across a shared stream.
Stream-GDPO optimizes complete multi-turn rollouts using separately normalized outcome, format, recall, and memory rewards.
It aligns training with stateful streaming decisions whose early actions shape later states.
WTI sets a new open-source state of the art, reaching 83.3\% on StreamingBench and 73.6\% on OVO-Bench while leading its Real-Time, Backward, and Forward groups.
\end{abstract}
 \section{Introduction}

Streaming video assistance requires a model to answer asynchronous questions as a video unfolds, using only the observed prefix and an evolving interaction state.
Recent Video Large Language Models (Video-LLMs) have advanced offline video question answering, but most assume access to a complete video or pre-collected clip before inference~\citep{mvbench2024,longvideobench2024,videomme2025}.
In streaming settings, frames arrive continuously and future frames are unavailable.
For each question, the model should respond as soon as the active visual context and retained history provide sufficient evidence; otherwise, it must first supplement its state by revisiting previously observed visual evidence or continuing to observe until the required evidence becomes available to the model.

Recent streaming Video-LLMs model response states or online decisions over observed prefixes~\citep{videollm_online2024,dispider2025,vispeak2025,livecc2025,streamo2026,streambridge2025,vst2026,livestar2025}.
Other work uses online cache and memory management for streaming inputs~\citep{streamingvlm2026,flashvstream2025,streamforest2025}, or visual-token and hierarchical compression for pre-collected videos~\citep{longvu2025,videochatflash2026}.
In causal streaming, however, the retained state is formed before subsequent questions are known; visual details excluded by forward-only compression therefore become unavailable when later questions reveal their relevance.
Further reasoning over that state cannot recover omitted evidence; opaque latent summaries also obscure what remains and when it was observed in the stream.

\begin{figure*}[t]
  \centering
  \includegraphics[width=\textwidth]{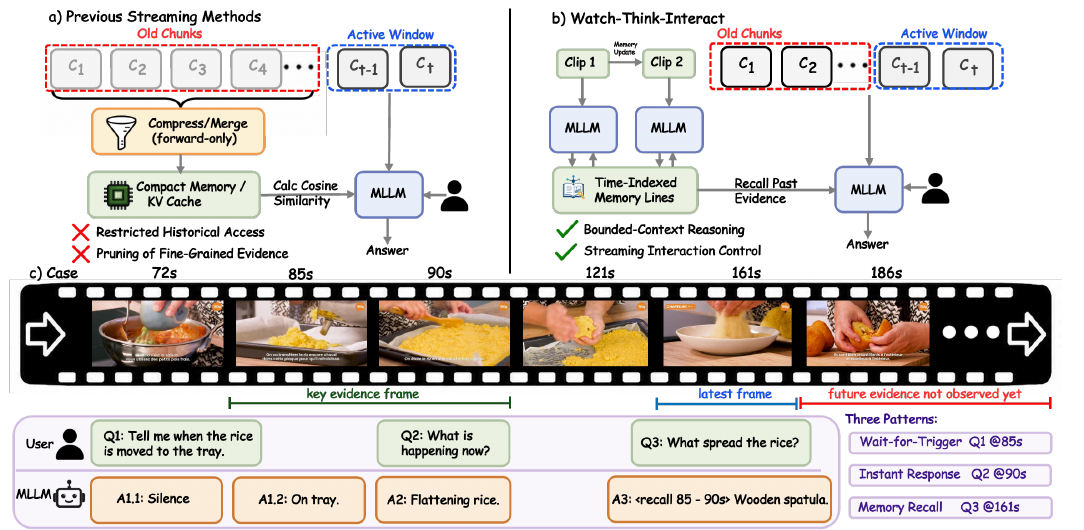}
  \caption{Motivation for Watch-Think-Interact. In causal streaming video, asynchronous questions may require current, past, or not-yet-observed evidence. WTI uses a bounded active window for current perception, compact time-indexed memory for direct reasoning and temporal localization, and selective recall to reload a past visual interval.}
  \label{fig:chronostream-motivation}
\end{figure*}

We therefore introduce Watch-Think-Interact (WTI), a closed-loop framework coupling active-window perception, compact time-indexed memory, response timing, and selective source-video recall.
Each memory entry pairs a semantic summary for direct reasoning with a source time range that localizes finer visual evidence when the summary is insufficient.
For every active question, a learned policy answers when the visible state suffices, remains silent when required evidence has not appeared, or recalls a relevant observed interval.
Recall returns evidence without advancing the stream, updates the visible state, and triggers a new decision at the same time step.
This action--feedback--redecision loop avoids replaying the full observed history while preserving on-demand access to source evidence.
Figure~\ref{fig:chronostream-motivation} illustrates how compact memory and selective recall reconnect the online state with source evidence.

\begin{figure*}[t]
  \centering
  \includegraphics[width=\textwidth]{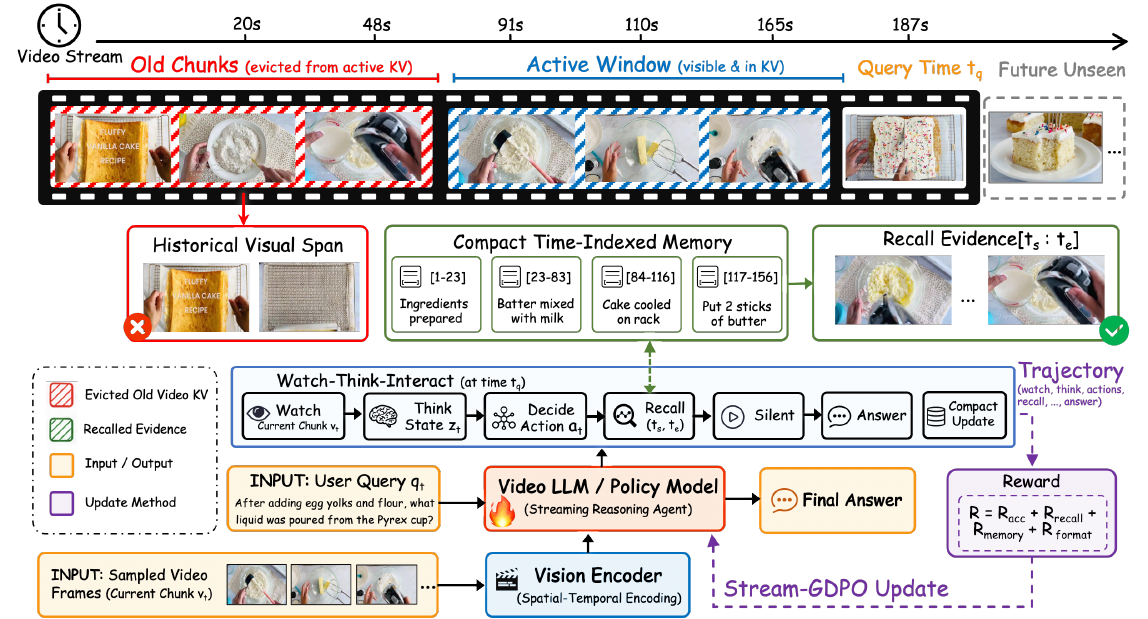}
  \caption{Overview of Watch-Think-Interact. For each arriving chunk, WTI updates a short thinking state and chooses to remain silent, answer, or recall. Recall returns visual evidence and loops back to the decision at the same time step; controller-triggered memory updates separately compress elapsed observations into a bounded temporal index. Stream-GDPO optimizes the resulting complete interaction trajectories.}
  \label{fig:method_overview}
\end{figure*}

To train this closed-loop behavior, we construct WTI-82K, which organizes 82,335 timed questions into 4,812 causally aligned multi-turn trajectories linking query times, answerable moments, supporting evidence, state updates, and interaction actions.
Relative to the 8-second active visual window, 56.9\% of trajectories span at least 120 seconds (15$\times$ the window), and 21.7\% span at least 240 seconds (30$\times$), so most trajectories extend far beyond the visible context.
Masked SFT initializes the interaction protocol, while Stream-GDPO extends GDPO~\citep{gdpo2026} into a trajectory-level reinforcement-learning objective for complete multi-turn rollouts, optimizing response timing, source-video recall, and memory updates throughout the stream.
WTI sets a new state of the art among open-source streaming models, reaching 83.3\% on StreamingBench~\citep{streamingbench2024} and 73.6\% weighted overall accuracy on OVO-Bench~\citep{ovobench2025}.

In summary, our contributions are as follows:
\begin{itemize}
  \item We introduce Watch-Think-Interact, a closed-loop framework for bounded multi-turn streaming reasoning. Its time-indexed memory and selective recall recover source evidence when later questions reveal its relevance.
  \item We construct WTI-82K, comprising 82,335 timed questions in 4,812 causally aligned trajectories. It supplies supervision for answering, waiting, recall, and memory updates across a shared stream.
  \item We develop Stream-GDPO for trajectory-level reinforcement learning over complete multi-turn rollouts. Its separate normalization of outcome, format, recall, and memory rewards aligns optimization with stateful streaming decisions throughout each rollout.
  \item WTI sets a new open-source state of the art on StreamingBench and OVO-Bench. It leads OVO-Bench's Real-Time, Backward, and Forward groups while retaining strong offline long-video performance.
\end{itemize}
 \section{Related Work}

\paragraph{Streaming Video Understanding.}
Video-LLMs have progressed from offline long-video reasoning, where the full video is available before inference, to online benchmarks requiring prefix-only answers, timestamped queries, and temporal multi-turn interaction~\citep{mvbench2024,longvideobench2024,videomme2025,ovobench2025,ovbench2025,svbench2025}.
Streaming systems add response-state modeling, feedback policies, proactive interaction, streaming thinking, or instruction tuning~\citep{videollm_online2024,dispider2025,streambridge2025,vst2026,streamo2026}.
These methods decide when to answer from the observed prefix; WTI addresses later turns whose source-linked evidence has left the active visual window.

\paragraph{Long-Stream Memory.}
Long-stream methods use visual-token or hierarchical compression, compact KV states, recurrent windows, and event memory to keep streaming context tractable~\citep{longvu2025,videochatflash2026,streamingvlm2026,flashvstream2025,streamforest2025}.
Their forward-only histories compress evidence before later questions arrive and provide no source-linked mechanism for reloading omitted visual details.
WTI instead makes compact memory time-indexed and source-recallable.

\paragraph{Agentic Multimodal Reasoning.}
Agentic multimodal models acquire evidence through tools or additional visual inspection~\citep{deepeyes2026, deepeyesv22026, gao2026videosearcher}, while memory agents learn to update or revisit compact state~\citep{memagent2026,rememr12026,wang2026explore}.
WTI brings both capabilities to streaming video by deciding whether to wait, answer, or recall time-bounded history while transferring compact state at controller-defined boundaries.
 \section{Method}

\begin{figure*}[t]
  \centering
  \includegraphics[width=\textwidth]{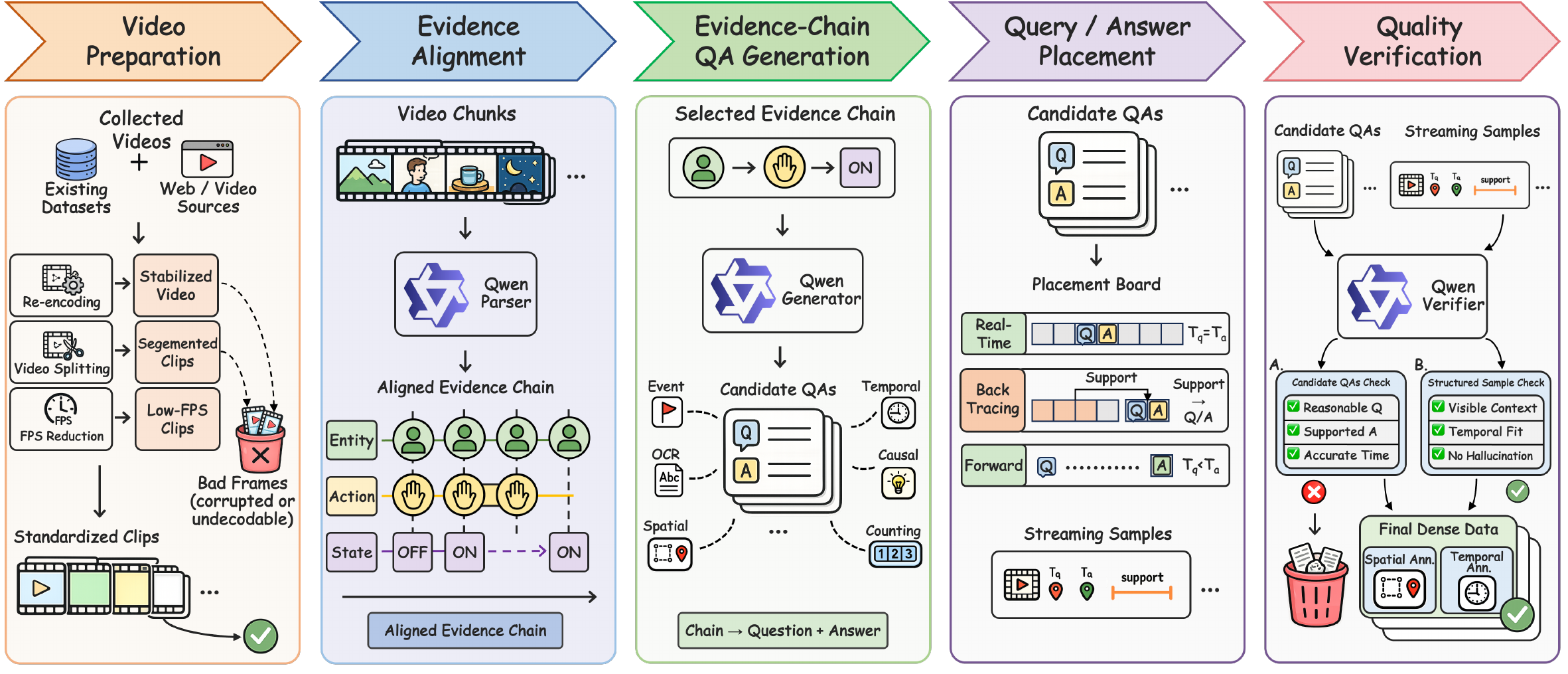}
  \caption{WTI-82K construction pipeline. WTI-82K converts videos into streaming-causal trajectories by standardizing clips, aligning chunk-level evidence chains, generating QAs from aligned evidence, placing query and answerable times, and checking timestamp bounds, answer availability, action grammar, and evidence-interval fields. The resulting samples cover Real-Time, Backward Tracing, and Proactive interactions under the same visibility constraints as inference.}
  \label{fig:data_pipeline}
\end{figure*}

\subsection{Problem Definition}

We study online multi-turn streaming video reasoning, where a video arrives as an ordered stream of chunks:
\begin{equation}
  V=(c_1,\ldots,c_T).
\end{equation}
At step \(t\), the observed prefix is \(c_{\le t}:=(c_1,\ldots,c_t)\); future chunks are unavailable.

The same stream may contain multiple questions arriving at different times:
\begin{equation}
  \mathcal Q=\{(q_r,\tau_r,\alpha_r)\}_{r=1}^{R}.
\end{equation}
Here \(\tau_r\) and \(\alpha_r\) are the question and answer timestamps used for data construction and evaluation, with \(1\le\tau_1\le\cdots\le\tau_R\) and \(\tau_r\le\alpha_r\le T\).
Although \(\alpha_r\), the time when sufficient visual evidence becomes available, is hidden at inference, the model must output one answer \(\hat y_r\) at \(\hat t_r\ge\tau_r\).
A response before \(\alpha_r\) violates streaming causality; otherwise it must be grounded in \(c_{\le \hat t_r}\).

\subsection{Watch-Think-Interact Framework}

We instantiate an online multimodal agent that receives chunk \(c_t\) at step \(t\) and acts at assistant subturn \(k\) from the bounded model-visible state
\begin{equation}
  s_{t,k}=(W_t,M_t,Q_t,H_t,E^{\mathrm{rec}}_{t,k}).
\end{equation}
Here \(W_t\) is the active visual window, \(M_t\) compact textual memory, \(Q_t\) the active question set, \(H_t\) interaction history, and \(E^{\mathrm{rec}}_{t,k}\) recalled evidence.
Time-indexed memory carries evidence boundaries beyond \(W_t\), while the observed prefix remains in a controller-managed source archive accessible only through recall.

Each assistant subturn combines a streaming thinking update with an interaction decision.
The model first emits a short update \(z_{t,k}\), then its learned policy selects \(a_{t,k}\sim\pi_\theta(\cdot\mid s_{t,k},z_{t,k})\) from
\begin{equation}
  \begin{aligned}
  a_{t,k}\in\mathcal A_{\mathrm{int}}
  =\{&\mathrm{silence},\mathrm{response}(q_r,y),\\
     &\mathrm{recall}(\tau_s,\tau_e)\}.
  \end{aligned}
\end{equation}
The actions \(\mathrm{silence}\) and \(\mathrm{response}\) respectively delay output and answer the current question, both terminating the current chunk; \(\mathrm{recall}\) requests earlier evidence without advancing the stream.

A recall interval \([\tau_s,\tau_e]\) uses absolute source-video seconds and is valid only when \(0\le\tau_s\le\tau_e<\gamma_t\), where \(\gamma_t\) is the start time of \(c_t\).
The environment adds the returned evidence to \(E^{\mathrm{rec}}_{t,k+1}\), after which the policy updates \(z_{t,k+1}\) and decides again at the same step \(t\).
Recall is therefore a non-terminal action--feedback--redecision loop.
Our protocol permits at most one recall invocation per current chunk, returning at most \(K_r=4\) observed chunks.

At a controller-triggered memory boundary \(b\), \(Z_b\) collects model-generated records since the previous boundary, and the model rewrites the pre-update memory:
\begin{equation}
  M_{\mathrm{post}}=\operatorname{Compact}_{\theta}(M_{\mathrm{pre}},Z_b).
\end{equation}
Under the fixed token budget \(\lVert M_{\mathrm{post}}\rVert_{\mathrm{tok}}\le B_M\), each line stores a concise semantic note and source time range.
The semantic note supports direct reasoning when sufficient, while the time range anchors selective inspection of finer source-video evidence.

WTI can process arbitrarily long streams without expanding the model-visible context: every decision uses a fixed active window, budgeted textual state, and at most four returned chunks, while the controller-managed source archive grows with the observed stream.

\subsection{Masked Supervised Fine-Tuning}

Masked SFT matches the causal, bounded inference state~\citep{streamingvlm2026}.
At step \(t\), its dense visual context is limited to the recent window
\begin{equation}
  W_t=\{c_{\max(1,t-K_v+1)},\ldots,c_t\},
\end{equation}
where \(K_v\) is the visual window size; we use \(K_v=8\) one-second chunks (16 frames).
Without exposing the full prefix, compact memory \(M_t\), active questions \(Q_t\), and teacher-forced interaction history \(H_t\) provide the remaining inference-time textual state.

We serialize stream inputs, assistant outputs, and environment observations temporally.
A stream-causal mask restricts assistant tokens to earlier steps and their causal prefix, while a label mask applies loss only to assistant-generated tokens; environment outputs remain conditioning context.
Because actions are sparse and state-update spans are longer, within each group we average token losses first per target type present in the trajectory and then across present types.
This yields \(\bar\ell_{\mathrm{act}}\) for silence, response, and recall targets and \(\bar\ell_{\mathrm{state}}\) for thinking and compact-memory targets.
The bucket-weighted SFT objective is
\begin{equation}
  \mathcal L_{\mathrm{SFT}}(\theta)
  =
  \lambda_{\mathrm{act}}\bar\ell_{\mathrm{act}}
  +
  \lambda_{\mathrm{state}}\bar\ell_{\mathrm{state}}.
\end{equation}
Here \(\lambda_{\mathrm{act}}\) and \(\lambda_{\mathrm{state}}\) balance sparse interaction-action targets against longer thinking and memory-update targets.
This stage cold-starts the interaction protocol in the inference-time causal state by supervising teacher action and state-update targets, while stream inputs, environment observations, and recall returns remain fixed conditioning context.

\begin{table*}[!t]
  \centering
  \small
\setlength{\tabcolsep}{3.25pt}
  \begin{tabular}{@{}lrrrrrrr|rrrr|rrrr|r@{}}
    \toprule
    \textbf{Model} &
    \multicolumn{7}{c|}{\textbf{Real-Time}} &
    \multicolumn{4}{c|}{\textbf{Backward}} &
    \multicolumn{4}{c|}{\textbf{Forward}} &
    \textbf{Overall} \\
    \cmidrule(lr){2-8}\cmidrule(lr){9-12}\cmidrule(lr){13-16}
    & \textbf{OCR} & \textbf{ACR} & \textbf{ATR} & \textbf{STU} &
    \textbf{FPD} & \textbf{OJR} & \textbf{Avg.} &
    \textbf{EPM} & \textbf{ASI} & \textbf{HLD} & \textbf{Avg.} &
    \textbf{REC} & \textbf{SSR} & \textbf{CRR} & \textbf{Avg.} &
    \textbf{Avg.} \\
    \midrule
    \multicolumn{17}{c}{\textit{Proprietary models}} \\
    \midrule
    GPT-4o                 & 69.8 & 64.2 & 71.6 & 51.1 & 70.3 & 59.8 & 64.5 & 57.9 & 75.7 & 48.7 & 60.8 & 27.6 & 73.2 & 59.4 & 53.4 & 59.5 \\
    Gemini-1.5-Pro         & 85.9 & 67.0 & 79.3 & 58.4 & 63.4 & 62.0 & 69.3 & 58.6 & 76.4 & 52.6 & 62.5 & 35.5 & 74.2 & 61.7 & 57.2 & 63.0 \\
    \midrule
    \multicolumn{17}{c}{\textit{Open-source offline models}} \\
    \midrule
    LongVU-7B              & 53.7 & 53.2 & 62.9 & 47.8 & 68.3 & 59.8 & 57.6 & 40.7 & 59.5 & 4.8  & 35.0 & 12.2 & 69.5 & 60.8 & 47.5 & 46.7 \\
    LLaVA-OV-7B            & 66.4 & 57.8 & 73.3 & 53.4 & 71.3 & 62.0 & 64.0 & 54.2 & 55.4 & 21.5 & 43.7 & 25.6 & 67.1 & 58.8 & 50.5 & 52.7 \\
    LLaVA-Video-7B         & 69.1 & 58.7 & 68.8 & 49.4 & 74.3 & 59.8 & 63.5 & 56.2 & 57.4 & 7.5  & 40.4 & 34.1 & 70.0 & 60.4 & 54.8 & 52.9 \\
    LLaVA-NeXT-Video-7B    & 69.8 & 59.6 & 66.4 & 50.6 & 72.3 & 61.4 & 63.3 & 51.2 & 64.2 & 9.7  & 41.7 & 34.1 & 67.6 & 60.8 & 54.2 & 53.1 \\
    Qwen2.5-VL-7B          & 67.8 & 55.1 & 67.2 & 42.1 & 66.3 & 60.9 & 58.9 & 51.5 & 58.8 & 32.2 & 47.5 & 34.1 & 67.6 & 60.8 & 63.6 & 57.3 \\
    Qwen3-VL-8B            & 75.2 & 58.7 & 72.4 & 57.3 & 70.3 & 59.2 & 64.8 & 56.6 & 69.6 & 38.7 & 54.4 & 38.8 & 67.6 & 52.5 & 63.5 & 61.4 \\
    \midrule
    \multicolumn{17}{c}{\textit{Open-source streaming models}} \\
    \midrule
    StreamForest-7B        & 68.5 & 53.2 & 71.6 & 47.8 & 65.4 & 60.9 & 61.2 & 58.9 & 64.9 & 32.3 & 52.0 & 32.8 & 70.6 & 57.1 & 53.5 & 55.6 \\
    Streamo-7B             & 77.2 & 66.1 & 76.7 & 45.5 & 66.3 & 72.8 & 67.4 & 55.6 & 58.1 & 33.9 & 49.2 & 30.8 & 57.6 & \textbf{82.5} & 57.0 & 57.9 \\
    VST-7B                 & 80.5 & 55.1 & 72.4 & 55.1 & \textbf{76.2} & 64.1 & 67.2 & 56.9 & 64.9 & 48.4 & 56.7 & 33.0 & 66.9 & 62.1 & 54.0 & 59.3 \\
    ViSpeak-7B             & 75.2 & 58.7 & 71.6 & 51.1 & 74.3 & 66.9 & 66.3 & 59.9 & 48.7 & 64.0 & 57.5 & 33.8 & 68.5 & 60.4 & 54.3 & 61.1 \\
    StreamBridge-7B        & 84.6 & 71.6 & 74.1 & 49.4 & 75.3 & 72.8 & 71.3 & \textbf{67.7} & 57.4 & 79.0 & 68.1 & 19.2 & 64.3 & 61.7 & 48.4 & 62.6 \\
    WTI-8B (Ours)          & \textbf{90.6} & \textbf{76.2} & \textbf{78.5} & \textbf{66.9} & \textbf{76.2} & \textbf{75.5} & \textbf{76.9} & 67.0 & \textbf{74.3} & \textbf{82.8} & \textbf{73.4} & \textbf{42.2} & \textbf{73.1} & 77.1 & \textbf{70.0} & \textbf{73.6} \\
    \bottomrule
  \end{tabular}
  \caption{Results on OVO-Bench. Scores are grouped by Real-Time, Backward, and Forward tasks; Avg. gives the corresponding group average, and Overall gives the benchmark average.}
  \label{tab:ovobench_main}
\end{table*}
 \begin{table*}[t]
  \centering
  \begin{minipage}[b]{0.62\textwidth}
    \vspace{0pt}
    \centering
    \small
    \setlength{\tabcolsep}{2.2pt}
    \begin{tabular}{@{}lrrrrrrrrrrr@{}}
      \toprule
      \textbf{Model} &
      \textbf{OP} & \textbf{CR} & \textbf{CS} & \textbf{ATP} &
      \textbf{EU} & \textbf{TR} & \textbf{PR} & \textbf{SU} &
      \textbf{ACP} & \textbf{CT} & \textbf{Avg.} \\
      \midrule
      \multicolumn{12}{c}{\textit{Proprietary models}} \\
      \midrule
      GPT-4o                 & 77.1 & 80.5 & 83.9 & 76.5 & 70.2 & 83.8 & 66.7 & 62.2 & 69.1 & 49.2 & 73.3 \\
      Gemini-1.5-Pro         & 79.0 & 80.5 & 83.5 & 79.7 & 80.0 & 84.7 & 77.8 & 64.2 & 72.0 & 48.7 & 75.7 \\
      \midrule
      \multicolumn{12}{c}{\textit{Open-source offline models}} \\
      \midrule
      LLaVA-OV-7B            & 80.4 & 74.2 & 76.0 & 80.7 & 72.7 & 71.7 & 67.6 & 65.5 & 65.7 & 45.1 & 71.1 \\
      Qwen2.5-VL-7B          & 77.9 & 76.6 & 78.6 & 80.9 & 76.7 & 77.0 & 80.6 & 65.5 & 65.7 & 52.9 & 73.3 \\
      Qwen3-VL-8B            & 79.9 & 77.3 & 81.1 & 84.3 & 76.7 & 77.9 & 79.6 & 71.1 & 69.6 & 46.3 & 75.2 \\
      MiniCPM-o-4.5-9B       & 80.9 & \textbf{88.3} & 82.0 & 85.0 & 78.9 & 85.4 & 78.7 & 67.1 & 75.6 & 56.0 & 78.2 \\
      \midrule
      \multicolumn{12}{c}{\textit{Open-source streaming models}} \\
      \midrule
      VideoLLM-online-8B     & 39.1 & 40.1 & 34.5 & 31.1 & 46.0 & 32.4 & 31.5 & 34.2 & 42.5 & 27.9 & 36.0 \\
      ViSpeak-7B             & 79.8 & 71.1 & 81.4 & 78.8 & 74.5 & 70.1 & 63.9 & 64.2 & 71.4 & 28.0 & 70.4 \\
      StreamBridge-7B        & 84.7 & 82.7 & 88.9 & 89.8 & 77.4 & 85.4 & \textbf{84.3} & 69.9 & 71.7 & 35.8 & 77.0 \\
      StreamForest-7B        & 83.1 & 82.8 & 82.7 & 84.3 & 77.5 & 78.2 & 76.9 & 69.1 & 75.6 & 54.4 & 77.3 \\
      VST-7B                 & 85.4 & 82.0 & 86.4 & 89.1 & 74.2 & 87.2 & 82.4 & 73.1 & 73.9 & 47.3 & 79.5 \\
      WTI-8B (Ours)          & \textbf{88.1} & 79.7 & \textbf{92.7} & \textbf{90.1} & \textbf{79.2} & \textbf{92.5} & 82.4 & \textbf{80.9} & \textbf{83.2} & 40.4 & \textbf{83.3} \\
      \bottomrule
    \end{tabular}
    \captionof{table}{StreamingBench real-time visual understanding results.}
    \label{tab:streamingbench_main}
  \end{minipage}
  \hfill
  \begin{minipage}[b]{0.36\textwidth}
    \vspace{0pt}
    \centering
    \small
    \setlength{\tabcolsep}{2.4pt}
    \begin{tabular}{@{}lrrr@{}}
      \toprule
      \textbf{Model} & \textbf{MLVU} & \textbf{V-MME} & \textbf{LVB} \\
      \midrule
      \multicolumn{4}{c}{\textit{Offline models}} \\
      \midrule
      LongVA-7B             & 56.3 & 52.6 & 56.3 \\
      LLaVA-OV-7B           & 64.7 & 58.2 & --   \\
      LongVU-7B             & 65.4 & 60.6 & --   \\
      Qwen3-VL-8B           & 66.7 & 63.5 & 60.7 \\
      \midrule
      \multicolumn{4}{c}{\textit{Streaming models}} \\
      \midrule
      StreamForest-7B       & \textbf{70.0} & 61.4 & -- \\
      StreamBridge-7B       & 69.6 & 64.4 & --   \\
      VST-7B                & --   & 64.9 & 58.0 \\
      WTI-8B (Ours)         & 69.9 & \textbf{66.9} & \textbf{62.9} \\
      \bottomrule
    \end{tabular}
    \captionof{table}{Long-video results.}
    \label{tab:offline_main}
  \end{minipage}
\end{table*}
 
\subsection{Stream-GDPO Reinforcement Learning}

Scalarizing multi-reward RL before normalization can let high-variance outcomes dominate weaker process signals.
GDPO instead normalizes each reward component within a sampled group before combining advantages~\citep{gdpo2026}; Stream-GDPO extends this principle to complete online rollouts in the inference-time chunk environment.
Each rollout contains one stream with one or more timed questions; policy actions use the current online state, while chunk arrivals and recall returns are environment feedback.
Thus, early answers, unnecessary silence, or missed recalls affect the states available to later decisions.

Stream-GDPO combines question-level outcome, format, and recall-use signals with one rollout-level memory signal.
For question \(q\) in rollout \(i\), the answer outcome is computed by a form-aware verifier:
\begin{equation}
  R_{\mathrm{out}}^{i,q}
  =
  \max_{y\in\mathcal Y_q}
  \mathbb I\!\left(\operatorname{match}(\hat y_{i,q},y)\right),
\end{equation}
where \(\mathcal Y_q\) is the valid-answer set; the verifier checks causal response timing, while \(R_{\mathrm{fmt}}^{i,q}\) validates whether the action follows the required streaming grammar.

For recall supervision, let \(\ell_q,u_{i,q}\in\{0,1\}\) indicate respectively whether the teacher recalls within question \(q\)'s decision interval and whether rollout \(i\) makes a syntactically and temporally valid recall there.
We gate the recall-tool reward by the answer outcome:
\begin{equation}
  R_{\mathrm{tool}}^{i,q}
  =
  \ell_q\,u_{i,q}\,R_{\mathrm{out}}^{i,q}.
\end{equation}
This rewards teacher-aligned recall only when followed by a correct, causally valid response.

For compact-memory writing, let \(S_{\mathrm{time}}(M)\) score time indexing and \(S_{\mathrm{keep}}(M)\) salient-fact retention.
For an update \(M_{\mathrm{pre}}^i\to M_{\mathrm{post}}^i\), its improvement reward is
\begin{equation}
  \begin{aligned}
  R_{\mathrm{mem}}^i
  =
  \frac{1}{2}\Big[
  &S_{\mathrm{time}}(M_{\mathrm{post}}^i)-S_{\mathrm{time}}(M_{\mathrm{pre}}^i)\\
  &+S_{\mathrm{keep}}(M_{\mathrm{post}}^i)-S_{\mathrm{keep}}(M_{\mathrm{pre}}^i)
  \Big].
  \end{aligned}
\end{equation}
Averaging over timed questions gives \(R_{\mathrm{out}}^i\), \(R_{\mathrm{fmt}}^i\), and \(R_{\mathrm{tool}}^i\), while \(R_{\mathrm{mem}}^i\) remains rollout-level because each memory update persists across subsequent decisions and can affect multiple later questions.
Let \(\mathcal C=\{\mathrm{out},\mathrm{fmt},\mathrm{tool},\mathrm{mem}\}\) with weights \(w_{\mathrm{out}}=w_{\mathrm{fmt}}=1\), \(w_{\mathrm{tool}}=\lambda_{\mathrm{tool}}\), and \(w_{\mathrm{mem}}=\lambda_{\mathrm{mem}}\).
The two \(\lambda\) coefficients control the relative strengths of recall-use and memory-quality feedback, while \(\epsilon_{\mathrm{norm}}>0\) below stabilizes normalization when a component has low group variance.
For each rollout group \(\mathcal G\), Stream-GDPO forms the trajectory advantage by component-wise normalization:
\begin{equation}
  A_i=\sum_{c\in\mathcal C} w_c
  \frac{R_c^i-\mu_c(\mathcal G)}
       {\sigma_c(\mathcal G)+\epsilon_{\mathrm{norm}}}.
\end{equation}
We assign \(A_i\) to all assistant action turns in the rollout, giving trajectory-level credit to silence/response, recall, and compact-memory decisions.
The policy is optimized with the clipped objective
\begin{equation}
  \mathcal J_{\mathrm{GDPO}}(\theta)
  =
  \mathbb E_{i,j}\!\left[
  \min(\rho_{i,j}A_i,\tilde\rho_{i,j}A_i)
  -\beta D_{i,j}
  \right].
\end{equation}
Here \(\rho_{i,j}\) is the action-turn probability ratio between \(\pi_\theta\) and \(\pi_{\theta_{\mathrm{old}}}\), \(\tilde\rho_{i,j}=\operatorname{clip}(\rho_{i,j},1-\epsilon_{\mathrm{low}},1+\epsilon_{\mathrm{high}})\), \(D_{i,j}\) is the reference-policy KL penalty at \(s_{i,j}\), and \(\beta\) controls its strength.
We minimize \(\mathcal L_{\mathrm{GDPO}}=-\mathcal J_{\mathrm{GDPO}}\), retaining trajectory-level credit without separate turn- or token-level advantages.

\section{WTI-82K: Causally Aligned Streaming Interaction Data}
\label{sec:data}

Prior online video benchmarks and streaming instruction datasets~\citep{ovobench2025,svbench2025,streambridge2025,streamo2026} define timestamped queries, multi-turn interaction, proactive responses, and streaming task annotations.
Building on this setting, we construct WTI-82K as stateful streaming interaction trajectories rather than isolated video-QA pairs.
It organizes 82,335 timed questions into 4,812 trajectories that align observed chunks, query times, answerable moments, supporting evidence, state updates, and interaction actions under the causal context available at each moment.
Relative to the 8-second active window, 56.9\% of trajectories span at least 120 seconds and 21.7\% span at least 240 seconds.
Each trajectory contains 17.1 questions on average, and 49.9\% of questions fall into historical, future, multi-state, or temporal-reasoning categories.
This yields repeated decisions over a shared evolving stream rather than isolated current-frame questions.
As shown in Figure~\ref{fig:data_pipeline}, our pipeline extracts temporal evidence, generates questions from aligned evidence, places each query and answerable moment on the streaming timeline, and checks timestamp bounds, answer availability, action grammar, and evidence-interval fields.

\noindent\textbf{Temporal evidence extraction.}
We first filter videos for clear visual content and temporally locatable events~\citep{gao2026tpru}.
For each retained video, we extract chunk-level evidence and link facts that persist or change across chunks.
This yields minimal supporting intervals for evidence-chain selection and time-anchored question generation.

\noindent\textbf{Candidate question generation.}
Given the aligned evidence, we use Qwen3.5-397B to generate questions and reference answers from local facts and cross-chunk evidence chains.
The candidates cover immediate perception, historical evidence, temporal progression, and proactive interaction, so each trajectory includes both current-prefix questions and questions requiring long-range state maintenance.

\noindent\textbf{Query-time and answer-time placement.}
Following timestamped online-video protocols~\citep{ovobench2025,svbench2025,streambridge2025}, we convert each candidate into a streaming sample by assigning query and answerable timestamps from its supporting evidence interval.
Real-Time samples are answerable at the query time from the current visible prefix.
Backward Tracing samples are also answerable at the query time, but their supporting evidence lies in previously observed chunks and therefore tests history maintenance and recall.
Proactive samples are not answerable when the query appears and require the model to wait until the earliest answerable moment, when decisive future evidence has arrived.
Automated checks cover temporal bounds, answer availability, action grammar, and evidence localization before final serialization.
The resulting set exposes answer-now, recall-before-answer, and wait-for-evidence behaviors under the visibility constraints used for evaluation.
The supplementary material reports the data statistics and question distribution in detail.
 \section{Experiments}
\label{sec:experiments}

\begin{figure*}[t]
  \centering
  \includegraphics[width=0.98\textwidth]{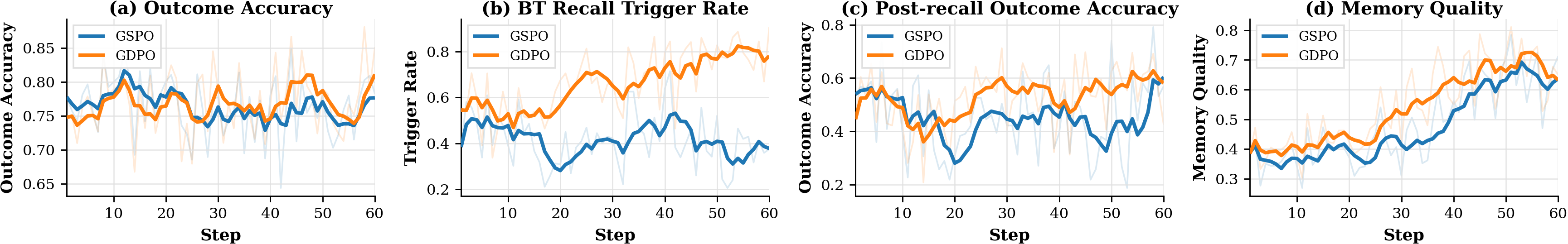}
  \caption{Training-time GSPO/Stream-GDPO diagnostics. Panels report outcome accuracy, BT recall triggering, post-recall outcome accuracy, and memory quality; faint and bold curves show raw and rolling values.}
  \label{fig:gspo_gdpo_step_comparison}

  \begin{minipage}[t]{0.62\textwidth}
    \vspace{0pt}
    \centering
    \small
    \setlength{\tabcolsep}{1.2pt}
    \begin{tabular}{@{}lrrrrrrrr@{}}
      \toprule
      \textbf{Config.} &
      \textbf{Stream.} &
      \textbf{OVO-RT} &
      \textbf{OVO-BT} &
      \textbf{OVO-FT} &
      \textbf{OVO Avg.} &
      \textbf{V-MME} &
      \textbf{MLVU} &
      \textbf{LVB} \\
      \midrule
      Base model & 75.2 & 64.8 & 54.4 & 63.5 & 61.4 & 63.5 & 66.7 & 60.7 \\
      + SFT      & 75.3 & 74.2 & 60.3 & 60.9 & 65.7 & 64.7 & 67.6 & 59.8 \\
      + GSPO     & 80.0 & 76.2 & 65.3 & 60.7 & 67.8 & 65.6 & 68.7 & 61.7 \\
      + GDPO     & \textbf{83.3} & \textbf{76.9} & \textbf{73.4} & \textbf{70.0} & \textbf{73.6} & \textbf{66.9} & \textbf{69.9} & \textbf{62.9} \\
      \bottomrule
    \end{tabular}
    \captionof{table}{Training-objective ablation across streaming and offline long-video benchmarks; GDPO denotes Stream-GDPO.}
    \label{tab:training_objective_ablation}
  \end{minipage}\hfill
  \begin{minipage}[t]{0.36\textwidth}
    \vspace{0pt}
    \centering
    \small
    \setlength{\tabcolsep}{1.2pt}
    \begin{tabular}{@{}lrrrr@{}}
      \toprule
      \textbf{Variant} & \textbf{OVO-BT} & \textbf{V-MME} &
      \textbf{MLVU} & \textbf{LVB} \\
      \midrule
      Full        & 73.4 & 66.9 & 69.9 & 62.9 \\
      w/o memory  & 62.8 & 63.5 & 44.7 & 61.6 \\
      w/o recall  & 55.5 & 48.3 & 54.2 & 57.9 \\
      memory only & 48.2 & 47.7 & 46.5 & 57.1 \\
      \bottomrule
    \end{tabular}
    \captionof{table}{Component ablation across OVO-BT and offline long-video benchmarks.}
    \label{tab:component_ablation_summary}
  \end{minipage}
\end{figure*}
 
\subsection{Experimental Setup}

\noindent \textbf{Implementation Details.}
WTI is initialized from Qwen3-VL-8B-Instruct~\citep{qwen3vl2025}.
Masked SFT trains on chunk-level trajectories with the vision encoder frozen and updates the language model and multimodal projector at a learning rate of $1\times10^{-5}$.
Stream-GDPO starts from the SFT checkpoint and uses veRL~\citep{sheng2024hybridflow} for one epoch on 8 H20 GPUs, with prompt batch size 64, 4 rollouts per prompt, actor learning rate $1\times10^{-5}$, asymmetric clipping \((\epsilon_{\mathrm{low}}=0.2,\epsilon_{\mathrm{high}}=0.28)\), and action-branch loss weight 0.15.
Videos are streamed as 1-second chunks at 2 FPS with an 8-chunk (16-frame) active window; recall is permitted at most once per current chunk and returns at most four observed chunks.
Compact-and-reprefill is triggered at the active-context boundary during inference.
SFT, RL, and evaluation share the same chunking, memory budget, recall tool, and action grammar; objective comparisons therefore do not change the streaming environment.

\noindent \textbf{Benchmarks.}
We evaluate OVO-Bench Real-Time, Backward, Forward, and weighted overall accuracy, followed by StreamingBench real-time visual understanding~\citep{ovobench2025,streamingbench2024}.
StreamingBench includes Object Perception (OP), Causal Reasoning (CR), Commonsense (CS), Action and Temporal Perception (ATP), Event Understanding (EU), Temporal Reasoning (TR), Proactive Reasoning (PR), Spatial Understanding (SU), Attribute and Character Perception (ACP), and Counting (CT).
We also report Video-MME~\citep{videomme2025}, MLVU~\citep{mlvu2024}, and LongVideoBench~\citep{longvideobench2024} official aggregate metrics for general long-video ability.

\noindent \textbf{Baselines.}
We compare with proprietary models, offline/static Video-LLMs, and online or streaming Video-LLMs, including GPT-4o~\citep{openai_gpt4o2024}, Gemini-1.5-Pro~\citep{gemini15_2024}, LongVA~\citep{longva2024}, LongVU~\citep{longvu2025}, LLaVA-series models~\citep{llavaonevision2024,llavavideo2024,llavanextvideo2024}, Qwen2.5-VL~\citep{qwen25vl2025}, Qwen3-VL~\citep{qwen3vl2025}, MiniCPM-o~\citep{minicpmo45_2026}, and recent streaming systems~\citep{videollm_online2024,vispeak2025,streambridge2025,streamforest2025,streamo2026,vst2026}.
All online evaluations are causal, and offline/static baselines are restricted to the observed prefix in online benchmarks under the benchmark protocol.

\subsection{Main Results}

Tables~\ref{tab:ovobench_main}--\ref{tab:offline_main} report OVO-Bench Real-Time, Backward, and Forward results, StreamingBench categories, and aggregate offline long-video performance.
On OVO-Bench, WTI achieves the best Overall, Real-Time, Backward, and Forward scores among open-source streaming baselines: 73.6\%, 76.9\%, 73.4\%, and 70.0\%, respectively; its overall score exceeds StreamBridge-7B (62.6\%) and ViSpeak-7B (61.1\%).
The gains span all three streaming modes: time-indexed memory and recall support historical reasoning, while active-window perception and multi-turn textual state retain Real-Time and Forward performance.

On StreamingBench, WTI reaches 83.3\%, ahead of VST-7B (79.5\%), StreamForest-7B (77.3\%), and StreamBridge-7B (77.0\%), and leads on OP, CS, ATP, EU, TR, SU, and ACP.
These category gains show that WTI's improvement extends from historical reasoning to current-prefix perception.
On offline long-video benchmarks, WTI retains strong general ability, scoring 66.9\% on Video-MME, 69.9\% on MLVU, and 62.9\% on LongVideoBench.
Among the reported streaming baselines, these are the best Video-MME and LongVideoBench scores; on MLVU, WTI remains within 0.1 point of StreamForest-7B (69.9\% vs.\ 70.0\%).
Together, these results establish state-of-the-art open-source streaming performance with strong offline long-video ability; the ablations below evaluate compact memory and source-video recall separately.

\subsection{Ablations}

Table~\ref{tab:training_objective_ablation} separates protocol initialization from trajectory-level policy optimization.
Masked SFT keeps StreamingBench at 75.3\% and raises OVO-RT from 64.8\% to 74.2\%, but reaches only 60.3\% on OVO-BT and 60.9\% on OVO-FT; its Forward score remains below the base model's 63.5\%.
Thus, imitation initializes recent-prefix behavior but remains substantially below the complete rollout-optimized model on Backward and Forward reasoning, motivating trajectory optimization in which early actions shape later states throughout the rollout.

Figure~\ref{fig:gspo_gdpo_step_comparison} compares the training dynamics of Stream-GDPO and GSPO~\citep{zheng2025gspo}.
Stream-GDPO sustains stronger BT recall triggering, post-recall accuracy, and memory quality, and raises OVO-BT from 65.3\% to 73.4\% and OVO-FT from 60.7\% to 70.0\%, lifting the overall score to 73.6\%.
It separately normalizes outcome, format, recall, and memory signals before aggregation.
These diagnostic gains extend beyond final-answer accuracy.

Table~\ref{tab:component_ablation_summary} shows that compact memory and recall are complementary rather than interchangeable.
Removing memory lowers OVO-BT from 73.4\% to 62.8\%, while removing recall lowers it to 55.5\%; the memory-only variant reaches only 48.2\%.
Offline, removing recall reduces Video-MME from 66.9\% to 48.3\%, while removing memory reduces MLVU from 69.9\% to 44.7\%.
The memory-only result supports this division of labor: compact text indexes history but cannot substitute for visual evidence, whereas recall restores details omitted by compression.
 \section{Conclusion}

We introduced Watch-Think-Interact (WTI), a causal framework for long-horizon multi-turn streaming video reasoning that combines bounded active-window perception, time-indexed memory, source-video recall, and response timing.
WTI-82K supplies causally aligned trajectories, while Stream-GDPO optimizes complete online rollouts in the same chunk-level environment used at inference.
WTI sets a new open-source state of the art on StreamingBench and OVO\mbox{-}Bench while retaining strong offline long-video performance; ablations verify complementary contributions from both memory and recall.
 
{\small

}

\end{document}